# Phrase Based Language Model for Statistical Machine Translation

Technical Report by

Chen, Geliang

# Abstract

Reordering is a challenge to machine translation (MT) systems. In MT, the widely used approach is to apply word based language model (LM) which considers the constituent units of a sentence as words. In speech recognition (SR), some phrase based LM have been proposed. However, those LMs are not necessarily suitable or optimal for reordering. We propose two phrase based LMs which considers the constituent units of a sentence as phrases. Experiments show that our phrase based LMs outperform the word based LM with the respect of perplexity and n-best list re-ranking.

Key words: machine translation, language model, phrase based

# Contents

# 1 Introduction

In the process of translation, reordering is a usual phenomenon. A LM is mainly used to reorder the sentences which were translated via the translation model.

Reordering generally occurs in phrase level. For example, when “小明前天打篮球” is translated to “Xiaoming played basketball the day before yesterday”, where “前天” is translated to “the day before yesterday” and “打篮球” is translated to “played basketball”, reordering occurs between “played basketball” and “the day before yesterday”.

However, the widely used word based LM is not necessarily optimal in this case. Also in the example above, in a bigram word based LM, the probability of “Xiaoming played basketball the day before yesterday” is

$$\begin{aligned}&\mathrm{P(Xiaoming\ played\ basketball\ the\ day\ before\ yesterday)}\\&\quad = \mathrm{P(Xiaoming \mid /start) * P(played \mid Xiaoming)}\\&\quad * \mathrm{P(basketball \mid played) * P(the \mid basketball) * P(day \mid the)}\\&\quad * \mathrm{P(before \mid day) * P(yesterday \mid before)}\end{aligned}$$

While the probability of “Xiaoming the day before yesterday played basketball” is

$$\begin{aligned}&\mathrm{P(Xiaoming\ the\ day\ before\ yesterday\ played\ basketball)}\\&\quad = \mathrm{P(Xiaoming \mid /start) * P(the \mid Xiaoming) * P(day \mid the)}\\&\quad * \mathrm{P(before \mid day) * P(yesterday \mid before) * P(played \mid yesterday)}\\&\quad * \mathrm{P(basketball \mid played)}\end{aligned}$$

Divide one probability by another:

$$\frac{\mathrm{P(Xiaoming\ the\ day\ before\ yesterday\ played\ basketball)}}{\mathrm{P(Xiaoming\ played\ basketball\ the\ day\ before\ yesterday)}} = \frac{\mathrm{P(the \mid Xiaoming) * P(played \mid yesterday)}}{\mathrm{P(played \mid Xiaoming) * P(the \mid basketball)}}$$

It is probably that the probability of the two sentences differs little in a word based LM, although they seem so different.

Some researchers have proposed their phrase based LM. Kuo and Reichl proposed a phrase based LM for SR which used an iteration to add new phrases in lexicon and to substitute the corpus with the new phrases, so as to reduce the word error rate (WER) and the perplexity.[1] Tang[2] used a similar method with Kuo and Reichl, they both used bigram count and unigram log likelihood difference as their measure function. The difference is that Tang also used mutual information and entropy as his measure function, while Kuo and Reichl used bigram log likelihood difference and correlation coefficient instead. Heeman and Damnati proposed a different LM which derived the phrase probabilities from a language model built at the lexical level and lowered the WER.[3]

Table 1 generalized their works. Unfortunately, these methods are not specifically developed for the MT application, and they did not consider reordering which is what we focus on and will not occur in SR application.

Table 1: Comparison between researchers' work

| Researcher | Kuo & Reichl | Tang | Heeman & Damnati |
|---|---|---|---|
| Area | SR | SR | SR |
| Content | LM using iteration to add phrases & substitute corpus with new phrases | LM using iteration to add phrases & substitute corpus with new phrases | LM in which probabilities of phrases are derived |
| Difference | use bigram log likelihood difference and correlation coefficient as measure function | use mutual information and entropy as his measure function | Phrase probability is derived |
| Result | WER & perplexity lower | Character & sentence accuracy higher, perplexity lower | WER lower |

In the rest of paper, we propose two phrase based LMs in which phrases are taken into account rather than words. We describe how these LMs are made up and what the probability and perplexity of a sentence should be in these LMs.

The experiments on IWSLT data show that our LMs outperform the standard word based LM with the respect of perplexity and n-best list reranking.

# 2 Review of the Word Based LM

## 2.1 Sentence probability

In standard word based LM, probability of a sentence is defined as the product of each word given its history. Probability of a sentence $w_1^m$ is

$$P(w_1^m) = \prod_{i=2}^{m} P(w_i \mid w_1^{i-1}) * P(w_1)$$

If we approximate $P(w_i \mid w_1^{i-1})$ to $P(w_i \mid w_{i-n+1}^{i-1})$ (i-n+1⩾1), we will have

$$P(w_1^m) \approx \prod_{i=n}^{m} P(w_i \mid w_{i-n+1}^{i-1}) * \prod_{i=2}^{n-1} P(w_i \mid w_1^{i-1}) * P(w_1)$$

This is the n-gram model.

## 2.2 Perplexity

A sentence's perplexity is defined as

$$PPL(w_1^m) = P(w_1^m)^{-\frac{1}{m}}$$

A text's perplexity is defined as

$$PPL(s_1^t) = (\prod_{i=1}^{t} P(s_i))^{-\frac{1}{N}}$$

where $s_i$ is the *i*-th sentence of the text and *N* is the total word number of $s_1^t$.

## 2.3 Smoothing

Generally, the probability of an n-gram is estimated as

$$P(w_i \mid w_{i-n+1}^{i-1}) = \frac{C(w_{i-n+1}^{i})}{C(w_{i-n+1}^{i-1})}$$

where $C(w_{i-n+1}^{i})$ is the count of $w_{i-n+1}^{i}$ that appeared in the corpus. But if $w_{i-n+1}^{i}$ is unseen, $P(w_i \mid w_{i-n+1}^{i-1})$ will be 0, so that any sentence that includes $w_{i-n+1}^{i}$ will be assigned probability 0.

To avoid this phenomenon, Good-Turing smoothing is introduced to adjust counts *r* to expected counts $r^*$ with formula

$$r^* = (r+1)\frac{N_{r+1}}{N_r}$$

where $N_r$ is the number of n-grams that occur exactly r times in corpus, and we define $N_0 = \sum_{i=1}^{\infty} i * N_i$.

Furthermore, a back-off model is introduced along with Good-Turing smoothing to deal with unseen n-grams:

$$P_{BO}\left(w_i \mid w_{i-n+1}^{i-1}\right) = \begin{cases} \alpha(w_i \mid w_{i-n+1}^{i-1}) & \text{if } C(w_{i-n+1}^{i}) > 0 \\ d\left(w_{i-n+1}^{i-1}\right) P_{BO}(w_i \mid w_{i-n+2}^{i-1}) & \text{else} \end{cases}$$

where

$$\alpha(w_i \mid w_{i-n+1}^{i-1}) = \frac{C^*(w_{i-n+1}^{i})}{C(w_{i-n+1}^{i-1})}$$

and

$$d\left(w_{i-n+1}^{i-1}\right) = 1 - \sum_{w_i} \alpha(w_i \mid w_{i-n+1}^{i-1})$$

where $C^*(w_{i-n+1}^{i})$ is the adjusted count of $w_{i-n+1}^{i}$ after Good-Turing smoothing.

# 3 Phrase Based LM

## 3.1 Model description

There are two phrase based LMs for us to propose. Both of them are based on probabilities of phrases, with the same estimation

$$P(p_i \mid p_{i-n+1}^{i-1}) = \frac{C(p_{i-n+1}^{i})}{C(p_{i-n+1}^{i-1})}$$

We consider only phrases that has at most *MPL* words, in our models, *MPL*=3. Given a sentence $w_1^m$, there are *K* segmentations $S_1^K$ that satisfy the *MPL* limit, and the *i*-th segmentation $S_i$ divides the sentence into $J_i$ phrases. In our models, we consider a single word also as a phrase.

### (1) Sentence probability

The probability of a sentence in the first model (sum model) is defined as

$$P(w_1^m) = \sum_{i=1}^{K} P(w_1^m, S_i) = \sum_{i=1}^{K} P\left(p_1^{J_i}, S_i\right) = \sum_{i=1}^{K} P\left(p_1^{J_i} \mid S_i\right) * P(S_i)$$

$$\approx \sum_{i=1}^{K} \prod_{j=1}^{J_i} P\left(p_j \mid p_{j-n+1}^{j-1}\right) * P(S_i)$$

where $P\left(p_j \mid p_{j-n+1}^{j-1}\right) = \begin{cases} P\left(p_j \mid p_{j-n+1}^{j-1}\right) & \text{if } j \geq n \\ P\left(p_j \mid p_1^{j-1}\right) & \text{if } 1 < j < n \\ P(p_1) & \text{if } j = 1 \end{cases}$ and $P(S_i) = \frac{1}{K}$.

The sentence probability formula of the second model (max model) is defined as

$$P(w_1^m) = \sum_{i=1}^{K} P(w_1^m, S_i) = \sum_{i=1}^{K} P\left(p_1^{J_i}, S_i\right) \approx P\left(p_1^{J_{i_0}}, S_i\right) = \prod_{j=1}^{J_{i_0}} P\left(p_j \mid p_{j-n+1}^{j-1}\right)$$

where

$$i_0 = \underset{i}{\operatorname{argmin}}\, PPL(w_1^m, S_i)$$

and $P\left(p_j \mid p_{j-n+1}^{j-1}\right)$ is same with that in sum model. The definition of $PPL(w_1^m, S_i)$ can be seen below.

## (2) Perplexity

Sentence perplexity and text perplexity in sum model use the same definition as that in word based LM.

Sentence perplexity in max model is defined as

$$PPL(w_1^m, S_i) = P(w_1^m, S_i)^{-\frac{1}{J_i}}$$

and

$$PPL(w_1^m) = P(w_1^m)^{-\frac{1}{J_{i_0}}}$$

where

$$i_0 = \underset{i}{\operatorname{argmin}}\, PPL(w_1^m, S_i)$$

Text perplexity in max model is defined as

$$PPL(s_1^t) = \left(\prod_{i=1}^{t} P(s_i)\right)^{-\frac{1}{N_0}}$$

where $N_0 = \sum_{j=1}^{t} \operatorname{argmax}_i P(w_1^m, S_i, s_j)$.

### (3) Smoothing

In phrase level, both models take back-off model along with Good-Turing smoothing, simply substituting $w_i^j$ to $p_i^j$ in the formulas. Moreover, we introduce an interpolation between phrase probability and product of single word probability:

$$P^*(p_j \mid p_{j-n+1}^{j-1}) = \lambda\ P(p_j \mid p_{j-n+1}^{j-1}) + (1-\lambda)\frac{\prod_{i=1}^{k} P(w_i)}{(\sum_{\text{single word } w} P(w))^k}$$

where phrase $p_j$ is made up of *k* words $w_1^k$. The idea of this interpolation is to make the probability of a phrase made up of *k* words smooth with a *k*-word unigram probability. In our experiments, $\lambda$ =0.43.

## 3.2 Algorithm of training the LM

Given a training corpus, our goal is to train a phrase based LM, i.e. to calculate $C(p_i^j)$ for all $p_i^j$ that $0 \le j - i \le \text{maxorder} - 1$. Therefore, for each sentence $w_1^m$, we should find out every k-grams that $0 \le k \le \text{maxorder} - 1$.

Any k-gram $p_i^{i+k-1}$ can be described with k+1 integers 0⩽b[0]<b[1]<⋯<b[k]⩽m, indicating that the first phrase is made up from word b[0]+1 to word b[1], the second phrase from b[1]+1 to b[2] … the k-th phrase from b[k-1]+1 to b[k], and $b[i] - b[i-1] \le MPL$ for all *i*. Moreover, any (k+1)-tuple satisfying the requests above corresponds with a $p_i^{i+k-1}$. Therefore, we only need to exhaust all the k-tuples satisfying the requests above, and that just takes an iteration procedure. The Algorithm is in Table 2.

Table 2: Algorithm of Training the LM
Input: training corpus $s_1^t$
Output: LM based on $s_1^t$

procedure main
for each sentence $w_1^m$ in $s_1^t$
    b[i]←0 for all i
    for b[0]=0 to m-1 do
        iter(1)
Use the n-gram counts to train LM

procedure iter(order)
if order<=maxorder then do all the things below
    for j=b[order-1]+1 to min(b[order-1]+MPL, n) do
        b[order]←j
        Output the *order*-gram corresponding with $b_0^{order}$
        iter(order+1)

## 3.3 Algorithm of calculating sentence probability and perplexity

Given a sentence *w* and phrase based LM (sum model or max model), it is easy to make an algorithm following the formula. The algorithms both for sum model and for max model are shown below in Table 3(1) and Table 3(2).

Table 3(1): Probability & Perplexity in sum model
Input: sentence $w_1^m$, the sum model
Output: probability & perplexity of $w_1^m$
sum←0
for all K segmentations of $w_1^m$:
    p←product of P*
    sum+=p
sum/=K
probability = sum
perplexity = $sum^{-1/m}$

Table 3(2): Probability & Perplexity in max model
Input: sentence $w_1^m$, the sum model
Output: probability & perplexity of $w_1^m$
max←0
for all K segmentations $S_i$ of $w_1^m$:
    p←product of P*
    if p>max {max←p; argmax←i}
probability←max
$m_0$←$J_{argmax}$
perplexity = $sum^{-1/m0}$

# 4 Experiments

We performed experiments using our phrase based models, both sum model and max model, on a large and a small data track. We evaluated performance by measuring perplexity and BLEU (Papineni et al., 2002)[4].

## 4.1 Task 1: Small Track IWSLT

We first report the experiments using our phrase based models on the IWSLT data (IWSLT, 2011). Because of the computational requirements, we only employed the models on sentences which contain no more than 8 words.

We took general word based LM described in Chapter 2 as a baseline method (Base). As shown in Table 4, the training corpus in English contains nearly 21 thousand sentences and 146 thousand words.

Table 4: Statistics of corpora in Task 1

| Data | Sentences. | Words | Vocabulary |
|---|---|---|---|
| Training | 20997 | 145918 | 11906 |
| Test | 1000 | 6965 | 1672 |

The resulting systems were evaluated on the test corpus, which contains 1000 sentences. We calculated the perplexities of the test corpus with different upper limits of order using both sum model and max model, with and without smoothing described in Chapter 3.

We show the results measured in perplexity only. As shown in Table 5, the perplexities in sum models, with and without smoothing, are lower than that in Base. The perplexities in max models are higher, probably because the formula of perplexity in max model is different.

Table 5: Perplexities of the test corpus in different models

| Limit | Word(Base) | Sum | Sum Smoo. | Max | Max Smoo. |
|---|---|---|---|---|---|
| Unigram | 287.04 | 67.89 | 89.05 | 475.47 | 705.11 |
| Bigram | 96.14 | 43.26 | 58.75 | 138.20 | 230.08 |
| Trigram | 89.91 | 43.33 | 58.94 | 125.60 | 210.14 |
| 4-gram | 90.39 | 43.42 | 59.02 | 127.24 | 212.55 |
| 5-gram | 90.90 | 43.44 | 59.04 | 128.20 | 214.16 |
| 6-gram | 90.98 | 43.45 | 59.04 | 128.50 | 214.56 |
| 7-gram | 91.00 | 43.45 | 59.04 | 128.75 | 215.07 |
| 8-gram | 91.01 | 43.45 | 59.04 | 128.67 | 214.85 |

## 4.2 Task 2: Large Track IWSLT

We evaluate our models on the IWSLT data using both models with and without smoothing. Also because of computational requirements, we only employed the models on sentences which contain no more than 15 words.

As shown in Table 6, the evaluations were done on Dev2010, on Tst2010 and on Tst2011 data. Because of computational requirements again, we only selected sentences which contain no more than 10 words, and we only considered 10 best translations of each sentence instead of 1000 bests. For convenience, we only list the statistics of the reference.

Table 6: Statistics of corpora in Task 2

| Data | Sentences | Words | Vocabulary |
|---|---|---|---|
| Training | 54887 | 576778 | 23350 |
| Dev2010 | 202 | 1887 | 636 |
| Tst2010 | 247 | 2170 | 617 |
| Tst2011 | 334 | 2916 | 765 |

The results are shown in Table 7. Max model along with smoothing outperforms the baseline method under all three sets. The BLEU score increases with 0.3 on Dev2010, 0.45 on Tst2010, and 0.22 on Tst2011.

Table 7: Performance in different models on three corpora

| Model | Dev2010 | Tst2010 | Tst2011 |
|---|---|---|---|
| Base | 11.26 | 13.10 | 15.05 |
| Word | 11.92 | 12.93 | 14.76 |
| Sum | 11.86 | 12.77 | 14.80 |
| Sum+Smoothing | 12.02 | 12.54 | 14.76 |
| Max | 11.61 | 12.99 | 15.34 |
| Max+Smoothing | 11.56 | 13.55 | 15.27 |

We compared the sentences which were chosen by max model with those chosen by baseline method. Table 8 shows two examples from the chosen sentences from the Tst2010 corpus. We list sentences chosen with the baseline method and in max model respectively, as well as the reference sentences. Our max model generates better selection results than the baseline method in these cases.

| Table 8: Sentence selection outputs with baseline method and in max model |
|---|
| (a) Baseline: but we need a success |
| Max model: but we need a way to success . |
| Reference: we certainly need one to succeed . |
| (b) Baseline: there 's a specific steps that |
| Max model: there 's a specific steps . |
| Reference: there 's step-by-step instructions on this . |

# 5 Conclusions

We showed that a phrase based LM can improve the performance of MT systems. We presented two phrase based models which consider phrases as the basic components of a sentence. By calculating the counts of phrases we can estimate the probabilities of phrases, and by segmenting the sentence into phrases we can calculate its probability and perplexity. The experiment results not only showed the models' outperforming, but also gave us confidence to improve them.

# 6 Acknowledgement

I hereby thank Dr. Jia Xu sincerely for guiding me in all the period of research for and writing of this thesis.

I must also thank my parents for their support and encouragement.

# References

[1] HKJ Kuo, W Reichl (1999), Phrase-Based Language Models for Speech Recognition. *Proceedings of EUROSPEECH, 1999*

[2]H Tang (2002), Building Phrase Based Language Model from Large Corpus. *ECE Master Thesis*

[3] PA Heeman, G Damnati (1997), Deriving Phrase-based Language Models. *Automatic Speech Recognition and Understanding, 1997. Proceedings., 1997 IEEE Workshop on*

[4] Papineni, K. A., S. Roukos, T.W., and W. J. Zhu. (2002), Bleu: a method for automatic evaluation of machine translation. *Proceedings of ACL, pages 311–318, Philadelphia, July.*

北京大学本科毕业论文审查表

| | | | | | |
|---|---|---|---|---|---|
| 学院（系） | 元培学院 | | | 专业 | 概率与统计 |
| 学生姓名 | [illegible] | | | 学号 | 00846107 |
| 导师姓名 | [illegible] | 职称 | [illegible] | 导师单位 | 清华大学 |
| 论文题目 | Phrase Based Language Model for Statistical Machine Translation | | | | |
| 一、选题来源 | 自拟 | 导师指定 | 从公布的选题目录中自选 | 导师的课题或项目 | 其他 |
| | | | | √ | |

| | |
|---|---|
| 二、中期检查（包括学生文献资料掌握情况、论文进展情况、以及存在的问题的检查） | [illegible] |
| 三、成绩评定（包括论文答辩和论文评阅两种方式，院系可据实际情况选择其一。答辩或评阅小组至少由两名成员组成。成绩按为优、良、中、及格、不及格评定） | 论文答辩（ ）论文评阅（√）选择请打"√"<br>成绩：优<br>答辩/评阅小组成员名单：[illegible]<br>答辩/评阅小组意见：[illegible]<br>组长签名：[illegible]<br>2013年6月[illegible]日 |

北京大学本科毕业论文导师评阅表

（成绩按优、良、中、及格、不及格评定）

| | | | | | |
|---|---|---|---|---|---|
| 学号 | 00846107 | 学生姓名 | [illegible] | 论文成绩 | 优 |
| 学院 | 元培学院 | | | 专业 | 概率与统计 |
| 导师姓名 | [illegible] | 导师单位 | 清华大学 | 职称 | [illegible] |
| 论文题目 | Phrase Based Language Model for Statistical Machine Translation | | | | |
| 导师评语（包含对论文的性质、难度、分量、综合训练等是否符合任务书要求的评价） | [illegible]<br>导师签名：[illegible] | | | | |